# Siamese Neural Encoders for Long-Term Indoor Localization with Mobile Devices

Saideep Tiku and Sudeep Pasricha
Department of Electrical and Computer Engineering
Colorado State University, Fort Collins, USA
{saideep, sudeep}@colostate.edu

*Abstract*–Fingerprinting-based indoor localization is an emerging application domain for enhanced positioning and tracking of people and assets within indoor locales. The superior pairing of ubiquitously available WiFi signals with computationally capable smartphones is set to revolutionize the area of indoor localization. However, the observed signal characteristics from independently maintained WiFi access points vary greatly over time. Moreover, some of the WiFi access points visible at the initial deployment phase may be replaced or removed over time. These factors are often ignored in indoor localization frameworks and cause gradual and catastrophic degradation of localization accuracy post-deployment (over weeks and months). To overcome these challenges, we propose a Siamese neural encoder-based framework that offers up to 40% reduction in degradation of localization accuracy over time compared to the state-of-the-art in the area, without requiring any re-training.

## I. INTRODUCTION

Contemporary geo-location services have eliminated the need for burdensome paper-based navigational maps that were dominant in the past. Owing to the localization technologies of today, our physical outdoor reality is now augmented by an additional layer of virtual map-based reality. Such a revolutionary shift has dramatically changed many aspects of human experience: geo-location data is now used for urban planning and development (roads, location of hospitals, telecom network design, etc.), augmented reality video games (Pokémon Go, Ingress Prime) and has even helped realize entirely new socio-cultural collaborations (Facebook marketplace, Meetup, etc.) [1].

Unfortunately, due to the limited permeability of GPS signals within indoor environments, such services cannot be easily extended into buildings such as malls, hospitals, schools, airports, etc. Indoor localization services can provide immense value, e.g., during emergency evacuations or when locating people indoors in need of critical medical attention. In the future, such services could inform the architects of building design and make augmented indoor living a reality. Towards this goal, indoor localization is experiencing a recent upsurge in interest [2], including from industry (e.g., Google [3], Apple [4]).

Although substantial progress has been made in this area (see Section II), recent works suggest fingerprinting-based indoor localization as the most favorable solution [2], [5]-[10]. While any form of radio fingerprinting works, the ubiquitous deployment of WiFi Access Points (APs), and the superior localization accuracies achieved through it make WiFi the clear choice of radio infrastructure to support indoor fingerprinting.

Conventionally, fingerprinting-based indoor localization consists of two phases. The <u>first</u> phase, known as the offline phase, comprises of capturing WiFi signal characteristics, such as RSSI (Received Signal Strength Indicator) at various indoor locations or Reference Points (RPs) in a building. The RSSI values from all APs observable at an indoor RP can be captured as a vector and represents a fingerprint associated with that RP. Such fingerprints collected across all RPs form a dataset, where each row in the dataset consists of an RSSI fingerprint along with its associated RP location. The collection of fingerprints to form the dataset is known to be a very time-consuming endeavor [11]. Consequently, publicly available datasets only contain a few fingerprints per RP (FPR). Using such datasets, a machine learning (ML) model can be trained and deployed on mobile devices (e.g., smartphones) equipped with WiFi transceivers. In the <u>second</u> phase, called the online phase, WiFi RSSI captured by a user is sent to the ML model running on the user-carried device, and used to compute and then update the user's location on a map of the indoor locale on the user's device display, in real time. Deploying such models on the user device instead of the cloud enables better data privacy, security, and faster response times [2].

Recent works report improved indoor localization accuracy through the use of deep learning-based classifiers [5]-[6]. This is attributed to their superior ability at discerning underlying patterns within fingerprints. Despite these improvements, factors such as human activity, signal interferences, changes to furniture and materials in the environment, and also removal or replacement of WiFi APs (in the online phase) introduce changes in the observed RSSI fingerprints over time that can degrade accuracy [8]-[10]. For instance, our experiments suggest that in frameworks designed to deliver mean indoor localization error of 0.25 meters, these factors degrade error to as much as 6 meters (section V.C.) over a short period of 8 months. Most prior efforts in the indoor localization domain often overlook the impact of such temporal variations during the design and deployment stages, leading to significant degradation of accuracy over time.

In this paper, we introduce *STONE*, a framework that delivers stable and long-term indoor localization with mobile devices, without any re-training. The main contributions of this work are:

- Performing an in-depth analysis on how indoor localization accuracy can vary across different levels of temporal granularity (hours, days, months, year);
- Adapting the Siamese triplet-loss centric neural encoders and proposing variation-aware fingerprint augmentation for robust fingerprinting-based indoor localization;
- Developing a floorplan-aware triplet selection algorithm that is crucial to the fast convergence and efficacy of our Siamese encoder-based approach;
- Exploring design tradeoffs and comparing *STONE* with state-of-the-art indoor localization frameworks.

## II. BACKGROUND AND RELATED WORK

Broadly approached, indoor localization methodologies can be classified into three categories: *(i)* static propagation model-based, *(ii)* triangulation/trilateration-based, and *(iii)* fingerprinting-based [2]. Static propagation modeling approaches depend on the correlation between distance and WiFi RSSI gain, e.g., [12]. These techniques are functionally limited to open indoor areas given that multipath or shadowing effects of signals attributed to walls and other indoor obstacles are not considered. These methods also required the cumbersome creation of a gain model for each individual AP. Triangulation/Trilateration-based methods use geometric properties such as the distance between multiple APs and the mobile device [13] (trilateration) or the angles at which signals from two or more APs are received [14] (triangulation). While such methodologies may be resistant to mobile device specific variability (device heterogeneity), they are not resilient to multipath and shadowing effects [6]. As discussed in Section I, WiFi fingerprinting-based approaches associate sampled locations (RPs) with the RSSI captured across several APs [5]-[10][23]-[25]. These techniques are known to be resilient to multi-path reflections and shadowing as the RP fingerprint captures the characteristics of these effects leading to more accurate localization than with the other two approaches.

Fingerprinting generally employs ML to associate WiFi RSSI captured in the online phase to the ones captured at the RPs in the offline phase [15]-[16]. Recent work on improving WiFi fingerprinting exploits the increasing computational capabilities of smartphones. For instance, Convolutional Neural Networks (CNNs) have been proposed to improve indoor localization accuracy on smartphones [5]-[6], [17]. One major concern with fingerprinting is the enormous effort required to manually collect fingerprints for training. Openly available fingerprint datasets often only consist of a few fingerprints per RP [10]. This motivates the critical need for indoor localization frameworks that are competitive with contemporary deep-learning-based frameworks but require fewer fingerprints to be deployed.

An emerging challenge for fingerprinting-based indoor localization (especially WiFi-based) arises from the fluctuations that occur over time in the RSSI values of APs [8]-[9], [18], [23]-[25]. Such temporal-variations in RSSI arise from the combination of many environmental factors, such as human movement, radio interference, changes in furniture or equipment placement, etc. This issue is further intensified when WiFi APs are removed or replaced by network administrators, changing the underlying fingerprint considerably [10]. This leads to a catastrophic loss in localization accuracy over time (discussed in section V.B).

The most straightforward approach to overcome temporal variation is to capture a large number of fingerprints over a long period of time (in offline phase). A deep-learning model trained using such a dataset would demonstrate resilience to degradation in localization accuracy as it witnesses (learns) the temporal fluctuations of RSSI values at various RPs. The work in [8] proposes such an approach by training an ensemble of models with fingerprints collected over a period of several hours. The authors then take a semi-supervised approach, where the models are refit over weeks using a mix of originally collected labeled fingerprints and pseudo-labeled fingerprints generated by the models. This process is repeated over several months to demonstrate the strength of this approach. However, the collection of fingerprints at a high granularity of RPs (small distance between RPs) over a long period of time in the offline phase is not scalable in practice.

To overcome the challenge of lack of available temporally diverse fingerprints per RP, the authors in [18] propose a few-shot learning approach that delivers reliable accuracy using a few fingerprints per RP. The contrastive loss-based approach prevents the model from overfitting to the training fingerprints used in the offline phase. Unfortunately, their approach is highly susceptible to long-term temporal variations and removal of APs in the online phase. This forces the authors to recalibrate or retrain their model using new fingerprints every month.

Attempting to achieve calibration-free indoor localization, some researchers propose the standardization of fingerprints into a temporal-variation resilient format. One such approach, known as GIFT [9], utilizes the difference between individual AP RSSI values to form a new fingerprint vector. However, instead of being associated with a specific RP, each GIFT fingerprint is associated with a specific user movement vector from one RP to another. However, GIFT degrades in accuracy over the long-term and is also highly susceptible to removal of APs (section V).

Considering the general stability of simple non-parametric approaches over the long term, such as K-Nearest-Neighbor (KNN), the authors in [21] propose Long-Term KNN (LT-KNN), which improves the performance of KNN in situations where several APs are removed. However, LT-KNN fails to deliver the superior accuracies promised by deep-learning approaches and needs to be re-trained on a regular basis.

*In summary*, most indoor localization solutions are simply unable to deliver stable localization accuracies over time. The few prior efforts that aim to achieve stable long-term localization either require large amounts of fingerprints per RP captured over time, or frequent re-training (refitting) of the model using newly collected fingerprints. Our proposed *STONE* framework provides a long-term fingerprinting-based indoor localization solution with lower overhead and superior accuracy than achieved by prior efforts in the domain, without requiring any re-training.

III. SIAMESE NETWORK AND TRIPLET LOSS: OVERVIEW

A Siamese network is a few-shot learning (requiring few labeled samples to train) neural architecture containing one or more identical networks [19]-[20]. Instead of the model learning to associate an input image with a fixed label (classification) through an entropy-based loss function, the model learns the similarity between two or more inputs. This prevents the model from overfitting to the relationship between a sample and its label. The loss function for a Siamese network is often a Euclidean-based loss that is either contrastive [19] or triplet [20].

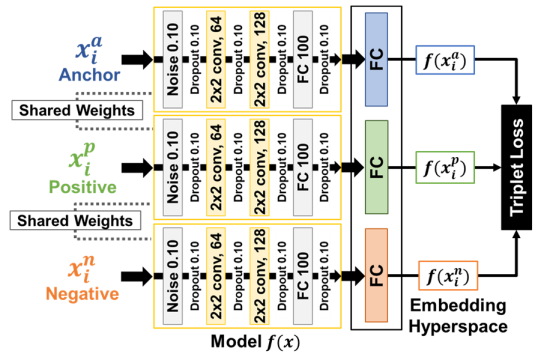

Figure 1: An example architecture of a Siamese encoder with triplet loss. A single CNN network is used, i.e., all the models share the same weights.

A Siamese network encoder using contrastive loss was proposed in DeepFace [19] for facial recognition. DeepFace focuses on encoding the input faces such that they are either pushed together or pulled apart in the embedded space based on whether they belong to the same person or not. The work in FaceNet [20] further improved on this idea using triplet loss that simultaneously pushes together and pulls apart faces of the same person and different persons, respectively.

An architectural representation of the Siamese model used in *STONE* (inspired by FaceNet) is presented in Fig. 1. The Siamese network consists of a single deep neural architecture. Note that given the specific model details (covered in section IV.D), the model itself can be treated as a black-box system.

The model in Fig. 1 can be represented as $f(x) \in R^d$ that embeds an image $x$ into a $d$-dimensional Euclidean embedding space. Therefore, the images $x_i^a$(anchor), $x_i^p$(positive) and $x_i^n$(negative) are embedded to form encodings $f(x_i^a), f(x_i^p)$ and $f(x_i^n)$ respectively, such that they belong in the same $d$-dimensional embedded hyperspace, i.e., $||f(x)||_2 = 1$. The anchor in a triplet is the reference label's sample with respect to which other label's samples are selected for the triplet. The triplet-based approach enables few-shot learning, as a single input to the training process is a combination of three different samples. Given a training set of $k$-classes and $n$-samples, the conventional classification approach [5]-[6], [15] has a total of $k \times n$ samples to learn from. In contrast, the triplet loss based approach has 3 samples per input, where each sample can be selected in $k \times n$ ways, i.e., a total of $(k \times n)^3$ inputs generated from the same dataset.

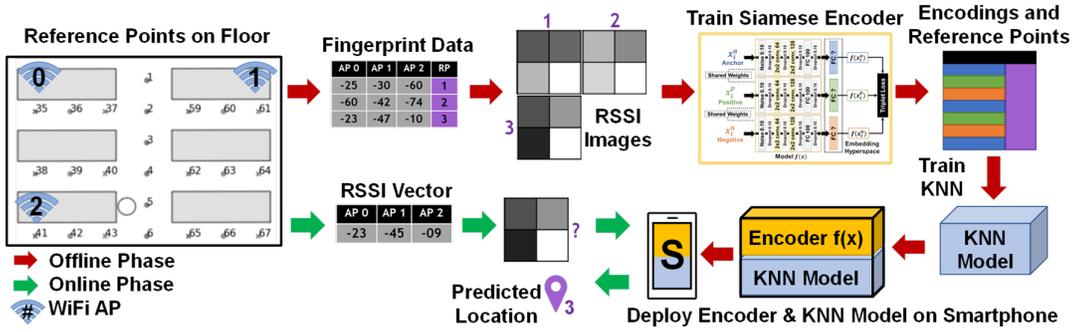

Figure 2: An overview of the STONE indoor localization framework depicting the offline (red arrows) and online (green arrows) phases.

The goal of the overall Siamese encoder is to ensure that the anchor image is closer to all other images of the same label (positives), than it is to any image of other labels (negatives). Based on this discussion, the embeddings should satisfy equation (1)

$$\|f(x_i^a) - f(x_i^p)\|_2^2 \leq \|f(x_i^a) - f(x_i^n)\|_2^2 \quad (1)$$

However, it is important to note that equation (1) can be trivially solved if $f(x) = 0$. Therefore, the margin $\alpha$ is introduced to enforce the stability of equation (1). Finally, the triplet loss function $L(x_i^a, x_i^p, x_i^n)$ that is to be minimized is given as:

$$L = \|f(x_i^a) - f(x_i^p)\|_2^2 - \|f(x_i^a) - f(x_i^n)\|_2^2 + \alpha \leq 0 \quad (2)$$

The authors of FaceNet [20] remark that to achieve rapid convergence it is important to select triplets that violate the constraint in eq. (1). Thus, for each triplet, we need to select a hard-positive $x_i^p$ that poses great dissimilarity with the anchor, and a hard-negative $x_i^n$ that poses great similarity with the anchor $x_i^a$. This may require the selection of triplets that satisfy both:

$$argmax_{x_i^p} \|f(x_i^a) - f(x_i^p)\|_2^2,$$
$$argmin_{x_i^n} \|f(x_i^a) - f(x_i^n)\|_2^2 \quad (3)$$

Evaluating $argmin$ and $argmax$ across the whole dataset is practically infeasible. To overcome this challenge, we present a novel and low-complexity indoor localization domain-specific approach for triplet selection in Section IV.

Once the embeddings for the training dataset have been produced, the embeddings and associated labels can be used to formulate a non-parametric model such as KNN. Later, this KNN model combined with the encoder can be used to classify an unlabeled sample as a known label.

Based on our discussion above, there are three salient features of Siamese networks that fit well to the challenges of long-term fingerprinting-based indoor localization: *(i)* Instead of associating a sample to its label, it learns the relationship between the samples of labels, *(ii)* Learning relationships between samples promotes generalization and suppresses the model's tendency to overfit the label-sample relationship, and *(iii)* It requires fewer samples per class/label to achieve good performance (few-shot leaning). Siamese networks will tend to avoid overfitting the training fingerprints and can minimize the offline fingerprint collection effort. The next section describes our framework that takes this approach for learning and classifying fingerprints.

## IV. STONE FRAMEWORK

### A. Overview

A high-level overview of the proposed framework is presented in Fig. 2. We begin in the offline phase (annotated by red arrows), where we capture RSSI fingerprints for various RPs across the floorplan. Each row in the fingerprint dataset consists of the RSSI values for each AP visible across the floorplan and its associated RP. These fingerprints are used to train the Siamese encoder depicted in Fig. 1. Once the Siamese encoder is trained, the encoder network itself is then used to embed the RSSI fingerprints in a $d$-dimensional hyperspace. The encoding of each RSSI vector and its associated RP, from the offline phase, form a new dataset. This new dataset is then used to train a non-parametric model. For our work, we chose the KNN classifier. At the end of the offline phase, the Siamese encoder and the KNN model are deployed on a mobile device.

In the online phase (green arrows), the user captures an RSSI fingerprint vector at an RP that is unknown. For any WiFi AP that is not observed in this phase, its RSSI value is assumed to be -100, ensuring consistent RSSI vector lengths across the phases. This fingerprint is pre-processed (see Section IV.B) and sent to the Siamese model. The encoding produced is then passed on to the KNN model, which finally predicts the user's location.

In the following subsections, we elaborate on the main components of the *STONE* framework shown in Fig. 2.

### B. RSSI Fingerprint Preprocessing

The RSSI for various WiFi APs along with their corresponding RPs are captured within a database as shown in Fig. 2. The RSSI values vary in the range of -100 to 0 dB, where -100 indicates no signal and 0 indicates a full (strongest) signal. The RSSI values captured are then normalized to a range of 0 (weakest) to 1 (strongest) signal. Finally, each RSSI vector is padded with zeros such that the length of the vector reaches its closest square. Each vector is then reshaped as a square image. This process is similar to the one covered by the authors in [6]. At this stage, in the offline phase, we have a database of fingerprint images and their associated RPs, as shown in Fig. 2.

### C. Long-Term Fingerprint Augmentation

A major challenge to maintain long-term stability for fingerprinting-based indoor localization is the removal of WiFi APs post-deployment (i.e., in the online phase) [10]. In the offline phase, it would be impossible to foretell which specific APs may be removed or replaced in the future. In the *STONE* framework, once an AP is removed or replaced, its RSSI value is set to -100. This translates into a pixel turning off in the input fingerprint image. *STONE* enables long-term support for such situations by emulating the removal of APs (turning off pixels of input images). When generating batches to train the Siamese encoder, we randomly set the value of a percentage of observable APs (*p_turn_off*) to 0. The value of *p_turn_off* is picked from a uniform distribution as described by:

$$p\_turn\_off = U(0.0, p\_upper) \quad (4)$$

where, *p_upper* is the highest percentage of visible APs that can be removed from a given fingerprint image. For the experiments in section VI, we chose an aggressive value of *p_upper=0.90*.

### D. Convolutional Neural Encoder

Given the superior pattern learning abilities of CNNs, we employ stacked convolutional layers to form the Siamese encoder. An architectural overview of the encoder is shown in Fig. 1. We use 2 convolutional layers (conv) with filter size of 2×2 with the stride set to 1 and consisting of 64 and 128 filters, respectively. They are followed by a fully connected (FC) layer of 100 units. The length of the embedding (encoder output or last layer) was empirically evaluated for each floorplan independently. Based on our analysis, we chose a value for this hyperparameter in the range of 3 to 10. To enhance the resilience of *STONE* to short-term RSSI fluctuations, Gaussian noise ($\sigma = 0.10$) is added to the model input (as shown in Fig. 1). Dropout layers are also interleaved between convolution layers to improve generalizability of the encoder. It is important to note that while the presented convolutional architecture functions well for our experiments and selected datasets, it may need slight modifications when porting to other datasets with a different feature space.

### E. Floorplan-aware Triplet Selection Algorithm

The choice of samples selected to form the triplets have a critical impact on the efficacy of the training and accuracy of the Siamese encoder. For a limited set of available fingerprints per RP (6-9 in our experiments), there are very few options in selecting a hard-positive. However, given an anchor fingerprint, selecting a hard-negative is a greater challenge due to the large number of candidate RPs across the floorplan. The motivation for our proposed triplet selection strategy is that RPs that are physically close to each other on the floorplan would have RSSI fingerprints that are the hardest to discern. This strategy is specific to the domain of fingerprinting-based indoor localization as the additional information of the relationship between different labels (location of labels with respect to each other) may not be available in other domains (such as when comparing faces).

To implement our hard-negative selection strategy, we first pick an RSSI fingerprint from an anchor RP, chosen at random. For the given anchor $RP_a$, we then select the negative $RP_n$ using a probability density function. Given the set of all *K* RPs, $\{RP_1, RP_2, ... RP_k\}$, the probability of selecting the $i^{th}$ RP as the hard-negative candidate is given by a bivariate Gaussian distribution around the anchor RP as described by the expression:

$$P(RP_i) \sim N_2(\mu_a, \sigma), \quad s.t. P(RP_a) = 0 \quad (5)$$

where $P(RP_i)$ is the probability of selecting it as the hard-negative and $N_2$ represents a bivariate Gaussian probability distribution that is centered around the mean at the anchor ($\mu_a$). However, another anchor fingerprint should never be chosen as the hard-negative, and therefore we set the probability of selecting an anchor to zero. The expression in (5) ensures that the RPs closest to the anchor RP have the highest probability of being sampled. This probability then drops out as we move away from the anchor. The bivariate distribution is chosen based on the assumption that the indoor environment under test is two-dimensional (a single floor). Once the anchor and the negative RPs are identified for a given triplet, the specific RSSI fingerprint for each is randomly chosen. This is because we have only a few fingerprints per RP, and so it is easy to cover every combination.

The proposed triplet selection strategy is subsequently used to train the Siamese model as discussed in Section IV.A, whose output is then used to train the KNN model in the offline phase.

In the online phase, the encoder and the KNN model are deployed on the mobile device and used to locate the user on the floorplan, as illustrated in Fig. 2 (lower half).

## V. EXPERIMENTS

### A. Experimental Setup

We evaluated the effectiveness of *STONE* across three large indoor paths derived from a publicly available dataset as well as based on our own measurement across multiple buildings. The next two subsections describe these paths, while the last subsection summarizes prior work that we compare against.

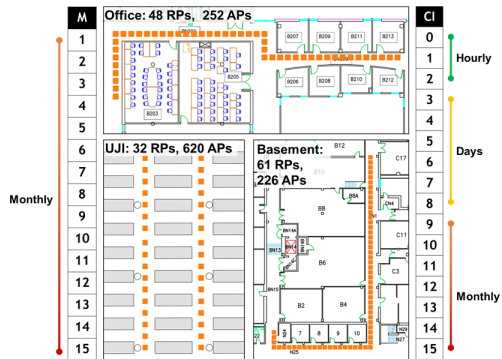

Figure 3: Indoor floorplans for long-term indoor localization evaluation, annotated with number of visible WiFi APs along the paths and RPs along the paths. Vertical scales show temporal granularities across months (left-UJI) and collection instances (right-Basement and Office).

#### 1) Fingerprinting Test Suite: UJI

*STONE* was evaluated on the public dataset UJI [10]. This dataset covers two floors within a library. However, due to high floorplan similarity across the two floors, we present the results for floor 3, for brevity. The dataset consists of fingerprints that are collected for the RPs along paths, with multiple fingerprints per RP that are collected at different instances of time. We utilize RPs from the dataset for which the fingerprints (up to 9) were collected on the same day for training the models we compared. The data from the following 15 months is used for testing. The UJI floorplan we considered is presented in Fig. 3 (bottom left of the figure). The RPs on the floorplan form a grid like structure over a wide-open area, which is different from the corridors evaluated for the Basement and Office indoor paths, discussed next.

#### 2) Fingerprinting Test Suite: Office and Basement

We also evaluated *STONE* at finer and broader granularity levels of hours, days, and months. The floorplan and associated details for these paths, captured from real buildings accessible to us, are presented in Fig. 3. The fingerprints were captured from two separate indoor spaces: Basement (61 meters in length) and Office (48 meters in length). An LG V20 mobile device was used to capture fingerprints along paths. While the Office path fingerprints are captured in a section of a building with newly constructed faculty offices, the basement path is surrounded by large labs that contain heavy metallic equipment. The Office and Basement paths are thus unique with respect to each other (and also the UJI path) in terms of environmental noise and multipath conditions associated with the paths. Each measured fingerprint location is annotated by an orange dot (Fig. 3) and measurements are made 1 meter apart. A total of 6 fingerprints were captured per RP at each collection instance (CI), under a span 30 seconds. The first three CIs (0–2), for both paths were on the same day, with each CI being 6 hours apart. The intention was to capture the effect of varying human activity across different times in the day; thus, the first CI is early in the morning (8 A.M), the second at mid-day (3 P.M), and the third is late at night (9 P.M). The following 6 CIs (3–8) were performed across 6 consecutive days. The remaining CIs (9–15) were performed on the following months, i.e., each was ≈30 days apart.

Fig. 4 depicts the ephemerality of WiFi APs on the Basement and Office paths across the 16 CIs (CIs:0–15) over a total span of 8 months. A black mark indicates that the specific WiFi AP (x-axis) was not observed on the indicated CI (y-axis). While capturing fingerprints across a duration of months, we did not observe a notable change in AP visibility up to CI:11. Beyond that, ≈20% of WiFi APs become unavailable. Note that the UJI dataset shows an even more significant change in visible WiFi APs of ≈50% around month 11; however, this change occurs much sooner in our paths, at C1:11, which corresponds to month 4 after the first fingerprint collection in CI:0. For the Office and Basement paths, we utilized a subset of CI:0 (fingerprints captured early in the morning) for the offline phase, i.e., training occurs only on this subset of data from CI:0. The rest of the data from CI:0 and CIs:1–15 was used for testing.

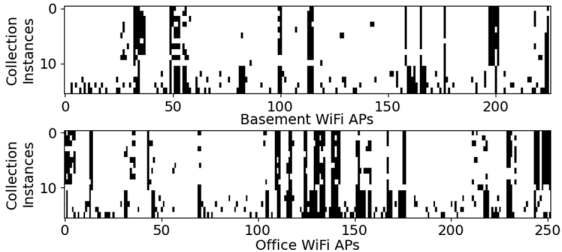

Figure 4: Ephemerality of WiFi APs across various collection instances for the Basement and the Office indoor paths.

*3) Comparison with Prior Work*

We identified four state-of-the-art prior works to compare against our proposed *STONE* framework. The first work, LearnLoc or KNN [11] is a lightweight non-parametric approach that employs a Euclidean distance-based metric to match fingerprints. The technique in the work is incognizant of temporal-variation and serves as a one of the motivations for our proposed work. The second work, LT-KNN [21], is similar to [11] but has enhancements to maintain localization performance as APs are removed or replaced over time. LT-KNN achieves this by imputing the RSSI values of APs that have been removed (are no longer observable on the floorplan) using regression. The KNN model is re-trained using the imputed data to maintain localization accuracy over time. The third work, GIFT [9], achieves temporal-variation resilience by matching the change in the gradient of WiFi RSSI values as the user moves along a path on the floorplan. Fingerprint vectors are used to represent the difference (gradient) between two consecutive WiFi scans and are associated with a movement vector in the floorplan. Lastly, the fourth work, SCNN [6], is a deep learning-based approach that has been designed to sustain stable localization accuracy in the presence of malicious AP spoofing. While SCNN is not designed to be temporally resilient, it is intended to maintain accuracy under the conditions of high RSSI variability. This makes SCNN an excellent candidate for our work to be compared against.

*B. Experimental Results: UJI*

Fig. 5 presents the mean localization error in meters (lower is better) for the proposed *STONE* framework and the four other prior fingerprinting-based indoor localization techniques across 15 months of the UJI dataset. Between months 1-2, we observe that most previous works (KNN, SCNN, LT-KNN) experience a sharp increase in localization error. Given that there is no temporal-variation in the training and testing fingerprints for month 1, previous works tend to overfit the training fingerprints, leading to poor generalization over time. In contrast, *STONE* remains stable and delivers ≈1 meter accuracy by not overfitting to the training fingerprints in month 1. We can also observe that GIFT provides the least temporal-resilience and has the highest localization error over time. The localization errors of *STONE*, SCNN, KNN and LT-KNN are around 2 meters (or less) up to month 10, followed by a severe degradation for KNN and SCNN. The significant change in APs at month 11, as discussed earlier, negatively impacts frameworks that are not designed to withstand the AP removal-based temporal-variation. In general, *STONE* outperforms all frameworks from months 2–11 with up to 30% improvement over the best performing prior work, LT-KNN, in month 9. Owing to the long-term fingerprint augmentation used in *STONE*, it remains stable and performs very similar to LT-KNN beyond month 11. Over the entire 15-month span, *STONE* achieves ≈0.3-meter better accuracy on average than LT-KNN. *Most importantly, LT-KNN requires re-training every month with newly collected (anonymous) fingerprint samples, whereas no re-training is required with STONE over the 15-month span.*

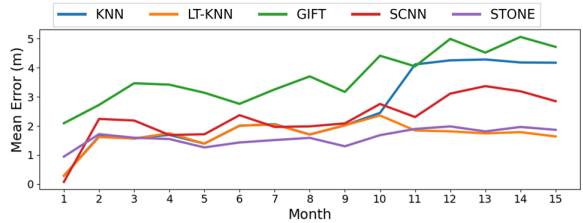

Figure 5: Comparison of localization error of various fingerprinting-based indoor localization frameworks over 15 months for the UJI indoor path.

*C. Experimental Results: Office and Basement*

Fig. 6 depicts the contrast in mean indoor localization errors across localization frameworks for the Office and Basement indoor paths. Similar to the previous results, most frameworks (especially SCNN and GIFT) tend to overfit the training fingerprints in CI:0 followed by a sharp increase in localization error for CI:1. It is worth noting that there is merely a difference of 6 hours between CI:0 and CI:1. In contrast to previous works, *STONE* undergoes the least increase in localization error initially (CI:0–1), followed by a fairly slow increase in localization error. We observe that across both indoor paths, GIFT and SCNN tend to perform the worst overall. While both these techniques show some resilience to temporal variation at the hourly (CIs:0–2) and the daily scale (CIs:3–8), they both tend to greatly lose their efficacy at the scale of months (CIs:9–15). GIFT's resilience to very short-term temporal variation is in consensus with the analysis conducted by its authors, as it is only evaluated over a period of few hours [9]. We also note that SCNN performs worse on the Office and Basement paths, as compared to with the UJI path (previous subsection). This may be due to the larger number of classes (RPs) in the Office and Basement paths. Both KNN and LT-KNN perform well (1–2 meters of localization error) on the Basement path. However, the localization error of KNN tends to increase in later CIs, particularly on the Office path. *STONE* outperforms LT-KNN across most collection instances, including up to and beyond CI:11. *STONE* delivers sub-meter of accuracies over a period of weeks and months and performs up to 40% better than the best-known prior work (LT-KNN) over a span of 24 hours (CI:1–3 in Fig. 5(b)), with superior localization performance even after 8 months. On average, over the 16 CI span, *STONE* achieves better accuracy than LT-KNN by ≈0.15 meter (Basement) and ≈0.25 meter (Office). As discussed earlier, *STONE* achieves this superior performance without requiring re-training, unlike LT-KNN which must be re-trained at every CI.

Overall, we attribute the superior temporal-variation resilience of *STONE*, to our floorplan-aware triplet selection, long-term AP augmentation, and also the nature of Siamese encoders that learn to differentiate between inputs instead of learning to classify a specific pattern as a label is also credited.

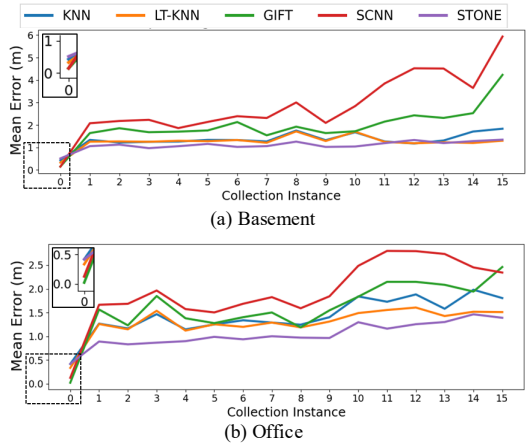

Figure 6: Localization errors of various frameworks over CIs for the Basement and Office indoor paths. Results for CI:0 are enlarged in the inset.

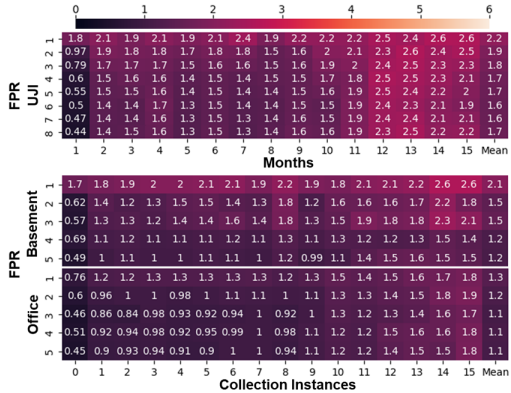

Figure 7: Sensitivity analysis on *STONEs* performance across varying number of fingerprints per RP (FPR) on UJI, Basement, and Office paths. Numbers in the heatmap cells show the obtained mean localization error.

### D. Results: Sensitivity to Fingerprints Per RP

Considering that *STONE* is explicitly designed to deliver the best temporal-resilience using minimal fingerprints, we performed a sensitivity analysis by varying the number of fingerprints per RP (FPR) across all indoor paths considered, to study its impact on localization error. Fig. 7 depicts the mean localization error as a heatmap (x-axis: timescale, y-axis: FPR) for different variants of *STONE*, each trained using a different number of FPRs. The final column in Fig. 7 represents the mean localization error across the timeline. The experiment is repeated 10 times with shuffled fingerprints to avoid any form of fingerprint selection bias. From the figure, we observe that for all three indoor paths, the *STONE* framework when trained using 1 FPR performs the worst; conversely increasing FPR beyond 4 does not produce notable improvements. Overall, these results show that *STONE* is able to produce competitive indoor localization accuracy in the presence of temporal-variations using as few as 4 FPR. To contrast this with a conventional classification-based approach, SCNN [6] is deployed using as many as 8 FPR (2×) and is unable to deliver competitive localization errors over time. Moreover, mobile devices can take several seconds to capture a single fingerprint (WiFi scan), thus reducing the number of FPRs in the training phase can save several hours of manual effort.

## VI. CONCLUSION

In this paper, we presented an effective temporal-variation resilient fingerprinting-based indoor localization framework called *STONE*. Our approach was evaluated against four state-of-the-art indoor localization frameworks across three distinct indoor paths. The experimental results indicate that *STONE* often delivers sub-meter localization accuracy and when compared to the best performing prior work, delivers up to 40% better accuracy over time, without requiring any re-training or model updating after the initial deployment. The ideas highlighted in this work, culminating in the *STONE* framework, represent promising directions for achieving low-overhead stable and long-term indoor localization with high-accuracy, while requiring the use of only a handful of fingerprints per reference point.

REFERENCES

[1] "20 Ways GIS Data is Used in Business and Everyday Life", 2016, *[online] https://nobelsystemsblog.com/gis-data-business/*
[2] C. Langlois, S. Tiku, S. Pasricha, "Indoor localization with smartphones", *6(4), IEEE Consumer Electronics,* 2017
[3] WiFi Location: Ranging with RTT, 2021, *[Online] https://developer.android.com/guide/topics/connectivity/wifi-rtt*
[4] Apple Indoor Maps, 2021, *[Online] https://register.apple.com/indoor*
[5] A. Mittal, S. Tiku, S. Pasricha, "Adapting Convolutional Networks for Indoor Localization with Smart Mobile Devices," *ACM GLSVLSI,* 2018
[6] S. Tiku, S. Pasricha, "Overcoming security vulnerabilities in deep learning-based indoor localization frameworks on mobile devices," *TECS, 18(6),* 2020
[7] L. Wang, S. Tiku, S. Pasricha, "CHISEL: Compression-Aware High-Accuracy Embedded Indoor Localization with Deep Learning", *IEEE Embedded System Letters,* 2021
[8] D. Li, et al., "Train Once, Locate Anytime for Anyone: Adversarial Learning based Wireless Localization," *INFOCOM,* 2021
[9] Y. Shu, et al., "Gradient-Based Fingerprinting for Indoor Localization and Tracking," *Transactions on Industrial Electronics, 63:4,* 2016
[10] M. Silva, "Long-Term WiFi Fingerprinting Dataset for Research on Robust Indoor Positioning," *MDPI Data*, 2018
[11] S. Pasricha, et al., "LearnLoc: A framework for smart indoor localization with embedded mobile devices, *CODES+ISSS,* 2015
[12] K. Chintalapudi, A. P. Iyer and V. N. Padmanabhan, "Indoor localization without the pain," *Mobile Computing Networks*, 2010
[13] J. Schmitz, et al., "Real-time indoor localization with TDOA and distributed software defined radio: demonstration abstract," *IPSN*, 2016
[14] E. Soltanaghaei, A. Kalyanaraman and K. Whitehouse, "Multi-path Triangulation: Decimeter-level WiFi Localization and Orientation with a Single Unaided Receiver," *MobiSys*, 2018
[15] P. Bahl and V. N. Padmanabhan, "RADAR: an in-building RF-based user location and tracking system," *INFOCOM* 2000
[16] Y. Shu, et al., "Magicol: Indoor Localization Using Pervasive Magnetic Field and Opportunistic WiFi Sensing," *JSAC, vol. 33:7,* 2015
[17] M. Abbas, et al., "WiDeep: WiFi-based Accurate and Robust Indoor Localization System using Deep Learning," *PerCom,* 2019
[18] A. Pandey, et al., "SELE: RSS Based Siamese Embedding Location Estimator for a Dynamic IoT Environment," *IoT Journal,* 2021
[19] Y. Taigman, et al., "DeepFace: Closing the Gap to Human-Level Performance in Face Verification," *CVPR,* 2014
[20] F. Schroff, D. Kalenichenko and J. Philbin, "FaceNet: A unified embedding for face recognition and clustering," *CVPR,* 2015
[21] R. Montoliu, et al., "A New Methodology for Long-Term Maintenance of WiFi Fingerprinting Radio Maps," *IPIN,* 2018
[22] S. Tiku, P. Kale, S. Pasricha, "QuickLoc: Adaptive Deep-Learning for Fast Indoor Localization with Mobile Devices", *ACM TCPS 5(4),* 2021
[23] S. Tiku, S. Pasricha, "PortLoc: A Portable Data-driven Indoor Localization Framework for Smartphones", *IEEE D&T, 36(5),* 2019
[24] S. Tiku, S. Pasricha, B. Notaros, Q. Han, "A Hidden Markov Model based Smartphone Heterogeneity Resilient Portable Indoor Localization Framework", *JSA, vol. 108,* 2020
[25] S. Tiku, S. Pasricha, B. Notaros, Q. Han, "SHERPA: A Lightweight Smartphone Heterogeneity Resilient Portable Indoor Localization Framework", *IEEE ICESS,* 2019